# Hybrid Optimized Deep Convolution Neural Network based Learning Model for Object Detection

**Venkata Beri**
**PRO2014002@iiita.ac.in**
**Indian Institute of Information Technology – Allahabad**

**Abstract**

Object identification is one of the most fundamental and difficult issues in computer vision. It aims to discover object instances in real pictures from a huge number of established categories. In recent years, deep learning-based object detection techniques that developed from computer vision have grabbed the public's interest. Object recognition methods based on deep learning frameworks have quickly become a popular way to interpret moving images acquired by various sensors. Due to its vast variety of applications for various computer vision tasks such as activity or event detection, content-based image retrieval, and scene understanding, academics have spent decades attempting to solve this problem. With this goal in mind, a unique deep learning classification technique is used to create an autonomous object detecting system. The noise destruction and normalising operations, which are carried out using gaussian filter and contrast normalisation techniques, respectively, are the first steps in the study activity. The pre-processed picture is next subjected to entropy-based segmentation algorithms, which separate the image's significant areas in order to distinguish between distinct occurrences. The classification challenge is completed by the suggested Hybrid Optimized Dense Convolutional Neural Network (HODCNN). The major goal of this framework is to aid in the precise recognition of distinct items from the gathered input frames. The suggested system's performance is assessed by comparing it to existing machine learning and deep learning methodologies. The experimental findings reveal that the suggested framework has a detection accuracy of 0.9864, which is greater than current techniques. As a result, the suggested object detection model outperforms other current methods.

## 1. Introduction

One of the most fundamental and difficult challenges in computer vision is object detection. As a result, in recent years, object detection has garnered increased attention in a number of real-time applications [1]. This is used to evaluate whether a given instance of a class is present in an image by producing a bounding box that overlaps the object instance and obtaining detection accuracy despite partial occlusions, posture, scale, lighting conditions, location, and camera position [2]. Most object detectors, on the other hand, have issues when objects, such as those in aerial photographs, are in different orientations or have a different layout from the ones used during training. When oriented objects are intensively disrupted, the difficulties grow more significant. Spatial feature aliasing arises as a result of this at the intersection of reception fields [3].

Object identification, scene understanding from pictures, activity recognition, and anomalous behaviour detection from video are all examples of computer vision applications that use feature extraction and learning algorithms as the initial step [17]. Feature augmentation, which provides features in varied orientations for detector training, is one option for oriented object detection [18]. This straightforward technique, on the other hand, is one alternative for detecting oriented objects. Define RoI transformers, which perform a spatial transformation to RoI while learning the parameters under the supervision of oriented bounding boxes [4]. The method of finding instances of a single class of things from an image or video is known as single-object class detection. Multi-class object detection [5] is the process of detecting many types of items in a picture or video.

Advanced machine learning algorithms [19] have grown in popularity in recent years, but the feature representational capacity based on the object detection approach has significant limitations that make it difficult to achieve high accuracy [20, 25]. In order to produce better results, machine learning algorithms should also improve feature selection and pre-processing of intrusion detection data. Using Multi-Tree methods, you may increase overall item detection and accuracy [29] by altering the quantity of training data and building up several decision trees [26]. Several machine learning methods, such as decision tree, random forest, KNN, and DNN, are utilised to form an ensemble model for improving overall object recognition and obtaining improved accuracy [22, 24]. The ensemble model's detection accuracy was awful when compared to deep learning approaches. Furthermore, the data analysis revealed that the quality of data characteristics is a key component in determining the detection effect. [6]

Because of their robust feature representation capabilities, deep learning-based algorithms have lately dominated the top accuracy benchmarks for numerous visual-recognition tasks [30]. Several deep learning-based object recognition algorithms have had tremendous success in natural scene images, thanks to this and other publicly accessible natural image datasets [21]. To put it another way, deep learning approaches are becoming better all the time, especially when it comes to computer vision [28] and image processing problems [27] like finding instances of semantic pictures of a specific class in digital photos [23]. Various optimization strategies for object identification and recognition have been developed as a result of advancements in the field of deep learning. [7]

The HO-DNCC, a hybrid optimised deep convolutional neural network for recognising objects in visual frames, is introduced in this study. The HO-DCNN, a hybrid optimised deep convolutional neural network for recognising human objects in visual frames, is tuned using an optimization technique. The HO-DCNN is optimised using an optimization method. An optimization strategy is used to optimise the system parameters of the HO-DCNN structure [8].

## 2. Literature Review

A tiny and slender architecture was provided by Winoto, A.S., Kristianus, M., et al. [11], which was afterwards compared to state-of-the-art models. The two models that was utilised to implement that architecture are Custom-Net and CustomNet2. To reduce computational complexity while maintaining accuracy and the capacity to compete with existing DCNN models, each of these models employs three convolutional blocks. These models will be trained using ImageNet, CIFAR 10, CIFAR 100, and other datasets. The results will be compared in terms of accuracy, complexity, size, processing time, and trainable parameter. In terms of inaccuracy, trainable parameters, and complexity, they determined that one of the models, CustomNet2, outperformed MobileNet, MobileNet-v2, Dense-Net, and NAS-Net-Mobile.

To compensate for the shortcomings of the floating-point-based quantization technique, Kim, S., and Kim, H., et al. [9] suggested a fixed-point-based quantization method tailored for embedded systems. While keeping the benefit of fixed-point-based quantization suited for HW implementation, precision floating-point may be accomplished utilising just around 20% of the number of bits for weight parameters. As a consequence, the proposed strategy can significantly contribute to the commercialization of deep learning techniques by enhancing the building of object detection HW accelerators for embedded platforms.

To categorise images into ten categories, Doon, R., Rawat, T.K., et al. [10] used a Convolutional neural network. As a baseline, the network was trained using the CIFAR-10 dataset. The Adam optimization technique for updating weights provides us with the maximum accuracy in terms of picture classification. Model overfitting was reduced using regularisation and dropout techniques. Using deep network architecture, they were able to achieve 90 percent accuracy on the training data and 87.57 percent accuracy on the test set.

J. Gao, A.P. French, and colleagues [12] created a process for creating synthetic pictures from field photographs. There were a total of 2271 synthetic images and 452 outdoor images used for training. They also developed a deep neural network for recognising C. sepium and sugar beets based on the micro YOLO architecture. They recommend determining anchor box sizes based on an application-specific dataset rather than utilising default values when building YOLO-based neural networks. The inclusion of synthetic images in the training phase improved the performance of the formed network in detecting C. sepium. They find that their network outperformed prior networks such as YOLOv3 in terms of speed and accuracy.

Y. Liu, L. Gross, and others [13] proposed the USPP CNN framework for generating segmentation on high-resolution remote sensing pictures. A significant contribution of that work was the analysis of the benefits of existing FCN-based models and the development of a novel model demonstrating that the encoder-decoder and spatial pyramid pooling module were two powerful tools that needed to be combined to take effect for building segmentation. The research employed two public building datasets: the Massachusetts and INRIA Aerial Image Labeling databases. According to the results, the recommended USPP model obtains high accuracy on these two datasets.

Weng, Y., Sun, Y., et al.[14] used the self-built small-scale data set training model to autonomously extract the capture pose characteristics, and the R-FCN model to extract the candidate frame for screening to achieve capture positioning and rough angle estimation based on the RGB information of the captured scene. The Angle-Net approach uses a precise estimation of the gripping angle to detect grasping. Using R-FCN to produce a small number of reliable-candidate gripping postures greatly speeds up grabbing detection. When compared to earlier techniques, it not only improves the algorithm's timeliness by restricting traversal search, but it also enhances the accuracy of attitude angle identification by applying cascading Angle-Net.

To conduct driven behaviour recognition, Zhang, C., Li, R., et al. [16] developed a novel Interwoven Convolutional Neural Network (Inter-CNN). That architecture can extract information from multi-stream inputs that record driver actions from diverse views (e.g., side video, side optical flow, front video, and front optical flow) and fuse the recovered features to conduct precise classification. They also created an ensemble system using a temporal voting mechanism, which lowered the possibilities of misclassification and enhanced accuracy. Experiments with 50 participants on a real-world dataset indicate that their method correctly categorises 9 types of driving behaviours with 73.97 percent accuracy and 5 aggregated behaviour classes with 81.66 percent accuracy.

To increase object recognition accuracy in large-area remote sensing photos, Hu, Y., Li, X., et al. [15] developed SUCNN, an effective object detection framework. Artificial composite samples were created using the specified sample update method. The sample update strategy considerably improved the second-stage model by merging background pictures with target items based on erroneous and missing detections. The quantitative comparison results for three item categories reveal that the recommended method for object detection in large-area remote sensing images was successful and improved.

## 3. Research Gap

Due to the vast number of concerns, it is vital to investigate a new approach for human object detection and focus our attention on the gaps in this research. The research gaps are crucial since they provide as a foundation for contemporary advancements in this sector.

- It can be shown that illumination changes, as well as the management of dynamic backgrounds, is an extensively focused research challenge that is addressed by all main soft computing methodologies.
- Although there have been some attempts with the development of deep learning-based graph neural networks, it remains an open subject for academics, with the management of picture noise and occlusions remaining a key research difficulty.
- Object identification and tracking using optimization algorithms for films have been presented, with the goal of improving pixel classification. These applications have been proven to have a higher success rate than typical tracking algorithms, but developing a general model to monitor objects of all forms and sizes is still a long way off.

- Threshold controls the accuracy and precision of object boundaries in any detection and tracking algorithm and is crucial for accurate categorization of pixels as foreground or background. Using a soft computing-based strategy, an appropriate threshold detection algorithm targeted at giving an ideal value for a range of images.
- One of the biggest drawbacks of all soft computing systems is the time it takes to arrive at a global optimal solution; no such study exists that compares the settling times of two or more soft computing-based algorithms.

All of the aforementioned challenges necessitate a thorough examination of principles from computer vision and optimization theory. The present study's authors chose a topic that has received little attention. The application of soft computing techniques combined with the usage of entropy-based concepts for object recognition and tracking stands out from the rest. Following a thorough examination of prior studies and a knowledge of the requirements of diverse applications, a number of areas have been identified where researchers might concentrate their efforts in order to conduct future study.

**4. Research Methodology**

The procedures in the suggested approach are outlined below, and the proposed strategy includes significant phases including pre-processing, segmentation, and object recognition. Pre-processing is the process of removing superfluous data from a frame. The gaussian filter and background subtraction are used in pre-processing to improve picture quality and remove noise from photos. A phase of contrast normalisation is also included in the pre-processing method. based on entropy. The input images are segmented to create foreground and background images. It aids in the division of an input picture into foreground and background areas, which may then be normalised independently to increase the contrast of particular regions. One of the most prevalent techniques of automatically segmenting images into two unique groups is to employ an entropy-based algorithm.

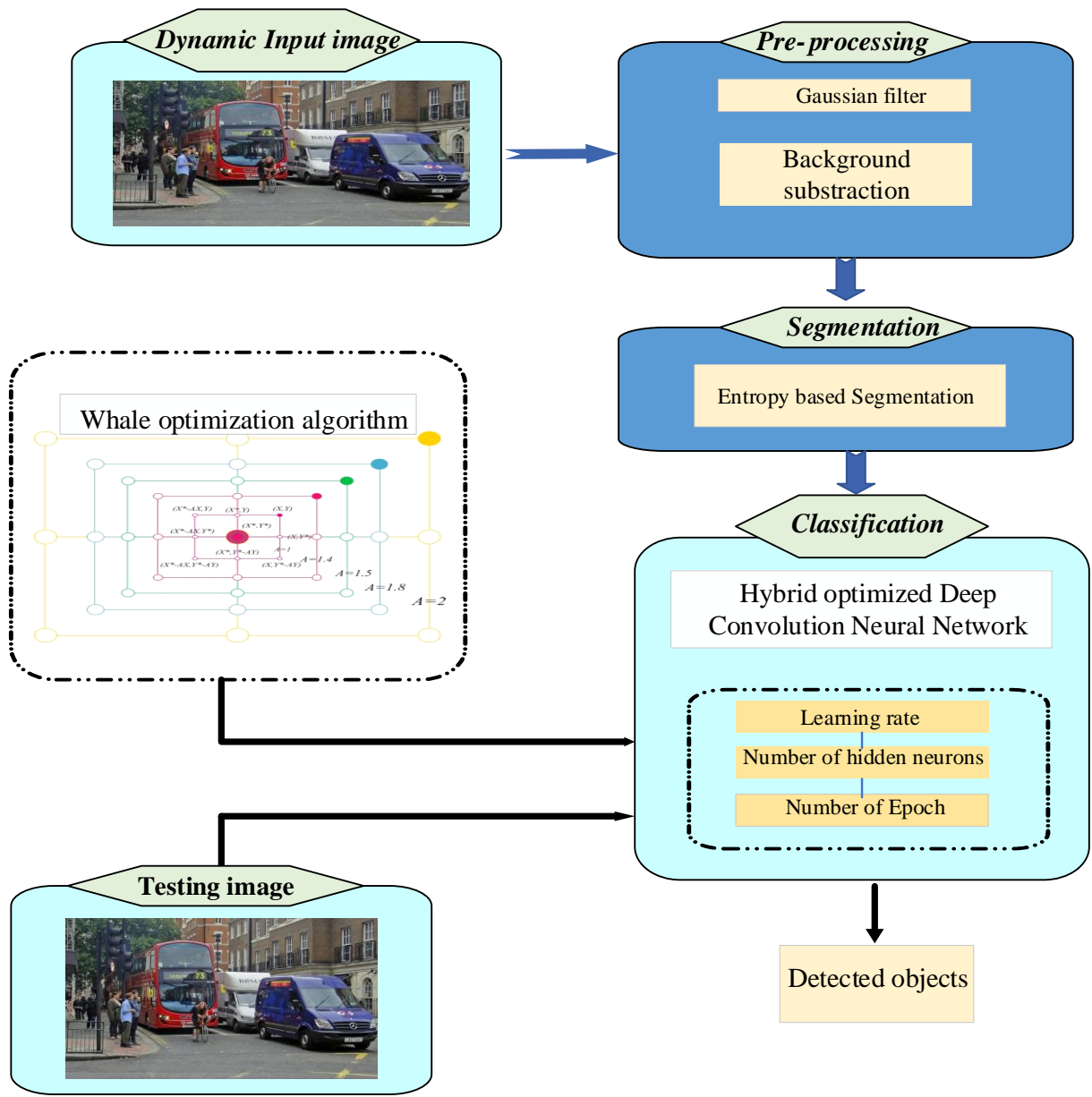

Figure 1: Layout of the proposed methodology

The suggested unique hybrid optimised deep convolutional neural network (HO-DCNN) is used to detect objects, and its parameters are customised using soft computing approaches such as whale optimization, horse herd optimization, black widow optimization, and rescue optimization models. It aids the suggested model in detecting the presence of an object in an image.

**4.1. Pre-processing procedures**

The input sizes of the dynamic image under consideration vary, therefore maintaining a consistent input size is essential. As a result, before applying the learning technique, the input images are shrunk to a predetermined pixel size. Therefore, pre-processing procedures are incorporated in the proposed methodology.

### 4.1.1. Gaussian Filtering

Typically, input source images are heavily contaminated with noisy information, necessitating the use of a noise removal process to enhance image quality. As a result, the scaled image $I(x,y)$ is combined with a Gaussian filter to smooth out the input image and reduce noise. The following equation can be used to express it mathematically.

$$G(x,y) = I(x,y) * \frac{e^{\frac{-(x^2+y^2)}{2\sigma^2}}}{\sum_x \sum_y e^{\frac{-(x^2+y^2)}{2\sigma^2}}} \qquad (1)$$

The sigma value is a two-dimensional quantity that describes the width of a gaussian function. The objects in the image are sharpened as a result of applying this filter. The filtered image is represented by $G(x,y)$. It also aids in the preservation of an image's edges, hence improving the image's quality.

### 4.1.2. Background subtraction

The goal of the background removal process is to determine the object's frontal regions in noiseless image frames. The majority of the films, as well as the changes in objects and individuals in the region of interest. As a result, object identification necessitates extracting and detecting them from a continual background. The background subtraction is expressed mathematically as follows:

$$I_k(y,z)_{b_s} = \{G(x,y) - b_r\} \qquad (2)$$

Where, $I_k(y,z)_{b_s}$ is the obtained subtraction procedure's with reference value of dissimilar time frames, $b_r$ is the background image with different time frames. While using background subtraction, the pixel intensity between two adjacent frames is kept constant when the backdrop is kept constant. Object detection is accomplished by comparing image pixel differences between the reference frame and the current video frame. The static pixels are ignored during the computation of variations, and these pixels reflect the identified changes.

### 4.1.3. Contrast Normalization for filtered image

In most cases, input images are captured using multiple scanners, and the brightness of the images may differ. As a result, images of varying intensity and contrast are produced, making operation and learning difficult. To overcome this obstacle, Contrast Normalization (CN) is employed to reduce the contrast of the deformed image. The filtered effect is now in the form of an RGB image that is channel-wise focused at the CN region. In the CN portion, the mean and standard deviations are determined for each data set. The provided value is extracted from the three filtered picture channels and separated by a standard deviation once both have been calculated. The following mathematical phrase is used to determine the CN class.

$$C(y,z) = \frac{I_k(y,z)_{b_s} - \mu}{\sigma} \quad (3)$$

Where $C(y,z)$ is the normalised image, $I_k(y,z)_{b_s}$ is the background subtracted image results, $\mu$ represents the estimated mean value for each channel, and $\sigma$ is the RGB channel's standard deviation. Equation 2 is applied to three channels to generate the normalised image. The proposed contrast normalisation phase improves the flexibility of the recommended model based on the required data values. CN also helps to reduce the impact of contrast variation on classification model training between images.

**4.2. Entropy based Segmentation**

Entropy is commonly used to calculate the ambiguity of a variable, which aids in the segmentation of the pre-processed picture. Consider the size $P * Q$ of an image frame. For $u \in \{1,2,...,P\}$ and $v \in \{1,2,...,Q\}$, the grey value of the pixel with coordinates $(u,v)$ may be represented as $f(u,v)$. Consider $K$ to be the sum of grey levels in image $I_j$, and $H$ to be the set of all grey levels $\{0,1,2,...,K-1\}$, in such a way that:

$$f(u,v) \in H \forall (u,v) \in image \quad (4)$$

The image's normalised histogram may be expressed as follows: as $G = \{g_0, g_1, ..., g_{K-1}\}$. Adopting multi-thresholding problem to the above equation and the obtained expression is given as follows,

$$G(T) = g_0(t_1), g_1(t_2), ..., g_{K-1}(t_{l-1}) \quad (5)$$

$$O_t = \max_T \{G(T)\} \quad (6)$$

Here $O_t$ symbolizes the optimal threshold function of shannons entropy. With the help of entropy-based segmentation, dissimilar objects present in a particular image taken for analysis is differentiated. Because the input here consists of numerous items in a single picture, it is utilised. As a result, we've used entropy-based segmentation to distinguish all of the items in a picture.

**4.3. Object Detection using Hybrid Optimized Deep Convolutional Neural Network**

Object identification has become increasingly important in research due to advancements in computer vision technology, and it is performed using a revolutionary deep learning technique. The suggested model is capable of extracting intrinsic characteristics in an automated manner while maintaining a high object identification rate. The traditional CNN structure is improved by keeping a large number of convolution layers. This, in turn, causes the network structure to improve the network's training time and learning capabilities. The reason for framing thick network layer is that elements such as form, size, and position incorporated in input pictures alter in different stages with its dynamic sensing behaviour, making them often changeable. As a result, obtaining the relevant class necessitates the extraction of a broader range of attributes. In order to achieve this goal, a dense collection of convolutional layers is used in the first layer to increase the network model's receptive field. By using a data sampling strategy on feature maps, this helps to compensate for the loss function that arises in the classification model. Convolution, pooling, rectified linear unit (RELU), batch normalisation, and classification layer comprising fully connected layer, logistic layer, and output classification layer are all included in the suggested Dense learning model. Thus, Physicians can categorise different things in an image in an effective manner using the suggested Dense CNN architecture without any prior information. The input layer of this network structure receives images and assigns biases and weight functions to them appropriately by modifying weight parameters. Training and testing are the two steps of this categorization methodology. Here, 80 percent of each image is connected with the training stage, while the remaining 20% is related with the testing stage.

*4.3.1. Convolution layer*

By using convolution between the input image and the filter, the convolutional layer aids in extracting features from the input image (kernel). The convolution process is carried out in this layer by carrying an element wise cross product of the input images and kernel, followed by a summing operation of these values. It produces output with a high amount of feature information that is low in dimension and invariant in feature space. The input grey scale image $(J)$ with size $(b \times c \times 1)$ is assigned with $(o)$ filter $(g)$ of size $(d \times e \times 1)$ and it then achieves a three-dimensional volume of feature map with size $\left(\frac{b-d}{stride}+1\right) \times \left(\frac{c-e}{stride}+1\right) \times (o)$. The stride function has a value of one, and the result must be represented as $(b-d+1) \times (c-e+1) \times (o)$. Simply, the stride indicates t how many steps a kernel makes each time. The following is the arithmetical formulation of the convolution operation among the filter vectors:

$$y_k^m = \varphi\left(\sum_d y_j^{m-1} * g_{de}^m + c_k^m\right) \qquad (7)$$

The activation function is represented as $\varphi$, where $m$ is the $m^{th}$ layer in the network design, and $g$ is the filter matrix (weight of the network). A tiny number of additional rows and columns (all zeros) are inserted to complete the convolution process at the border, known as zero padding. As a result, zero-pixel borders are applied to the image, making it the same size as the input. The quantity of feature information increases as the model structure grows, increasing the computational complexity and making the model more sensitive. In the following layer, a sampling technique is provided to solve this problem.

### 4.3.2. Pooling layer

By taking the output of the convolution layer, this layer aids in reducing the number of parameters from bigger image size. Downsampling is used at this phase to lower the dimensionality of each feature map. Despite the fact that this layer reduces the dimensionality of images, it maintains the important information. Subsampling is another term for it. Max pooling is always regarded the pooling filter in CNN, and it outsources the maximum value of the moving window over a two-dimensional input field. The output of the convolution layer $(b-d+1) \times (c-e+1) \times (o)$ is used with the $(f \times g \times 1)$ filter size to reduce the image dimension size, and it may be followed by $\left(\frac{(b-d+1)-f}{stride}+1\right) \times \left(\frac{(c-e+1)-g}{stride}+1\right) \times (o)$. As a result, image depth will be kept although dimension is lowered. The mathematical representation of this pooling layer is expressed as follows,

$$a_k^m = \varphi\left(\beta_k^m down(a_j^{m-1} + c_k^m)\right) \qquad (8)$$

A linear output is generated by each and every neuron. When a neuron receives this linear output, it responds with another linear output. Non-linear activation functions are used to overcome this and is expressed in the upcoming section.

### 4.3.3. Activation Function and Batch Normalization

The MRELU (Modified Rectified Linear Unit) function is used to learn the complex pattern. This layer's input value that is less than zero is set to zero; otherwise, the value is not changed. This layer enables the model to learn and perform more effectively.

$$g(a) = f(x) = \begin{cases} 0, & a < 0 \\ a, & a \geq 0 \end{cases} \quad (9)$$

The major role of Batch Normalization layer is to perform normalisation on each input channel. This batch normalisation layer speeds up deep network training and reduces the number of training epochs necessary.

### 4.3.4. Fully connected layer

The output of the pooling layer is flattened into a 1D feature vector and fed into a CNN architecture fully connected layer. Every neuron in this system is linked to other neurons, allowing the choice to be made based on the entire vision. For solving high demand object detection challenges, output layer introduces the cross-entropy loss. The result of the Logistic function is used to assign each input to one of the $C$ class labels, which is done using the cross-entropy loss function. The following equation shows the mathematical notation.

$$L(g) = \frac{1}{o}\sum_{w=1}^{o}\sum_{x=1}^{C}[t_{wx} \ln P(w = x) + (1 - t_{wx}) \ln(1 - P(w = x))] \quad (10)$$

As previously stated, $g$ denotes the weight, which is the convolution and fully linked layer's filter matrix. The total number of training samples is denoted by the letter $o$. The training sample and class indexes are $w$ and $x$, respectively. If the $w^{th}$ sample corresponds to the $x^{th}$ class, the output function $t_{wx} = 1\ (object\ detected)$, is used; otherwise, the output function $t_{wx} = 0\ (object\ not\ detected)$. Here $P(w = x)$ is the probability that the input belongs to the $x^{th}$ class predicted by the model. It's a function of the $g$ parameter. As a result, the loss function uses $g$ as a parameter. The optimization model's parameter tuning aids in determining the value of $g$ that minimises the loss function $L$.

The loss function in a deep learning model represents the network's prediction error. The optimization strategy, in particular for parameter adjustment in CNN, offers promising results by lowering the loss function. Kernel size, feature map count, and pooling type are all factors in CNN. The stride parameter is not optimised in this case because if it is, the image size shrinks to the point where CNN can no longer be applied. The setting of parameter values is a major concern in classifier algorithms. Using the Fitness Dependent Whale Optimization technique, CNN parameters are tweaked to minimise the loss function, i.e. predicted error.

*4.3.5. Training model*

The suggested whale optimization approach has a substantially higher convergence rate than previous optimization strategies. So that we may attain the lowest possible error while maintaining the highest level of precision. The parameters of DCNN are learned using an optimization approach while training the suggested classification network structure. The initialization phase is carried out for the purpose of training the classifier and is created by creating the initial solution randomly. This implies that parameters such as kernel size ($K_s$), feature map count ($F_m$), and pooling type ($P_t$) are set to the whale population. The following is the random value of these parameters in the search space:

$$I_u = \{K_{s=1\ to\ h} || F_{m=1\ to\ h} || P_{t=1\ to\ h}\} \quad (11)$$

$I_u$ specifies the number of interconnection layers to be optimised, whilst $h$ denotes the DCNN parameters initialization of whale. After the initialization procedure is completed, the fitness computation is assessed for reaching the suggested DCNN model's minimal loss function. The error function acquired by individual iteration determines that the hyper-tuning parameters are restored as a group. The following mathematical equation expresses as follows.

$$F_{(I_u)} = \min(L(g)) \quad (12)$$

The goal of this function is to use a meta-heuristic technique to discover the optimal DCNN settings. $L(g)$ refers to the DCNN's loss function; when the loss function is the smallest, the parameters of the associated network layer are highly suggested for classification and the update cycle continues. The optimization technique used in this study was motivated by whale hunting behaviour. To conduct the categorization action, this process follows three principles: searching, encircling, and attacking.

The social behaviour of these whales is observed in order to conclude the optimum search space, which aids in obtaining the best solution for the parameters used. The best solution produced is used as the aim for training the network model. Once the best search solution has been identified, the remaining search agents will seek to update their individual solutions. The goal of this optimization approach is that whales pursue prey in an encircling pattern for food, and the distance between whale and prey is mathematically stated by the following equation.

$$E = |F.q\vartheta^*(ju) - q\vartheta(ju)| \qquad (13)$$

$$q\vartheta(ju+1) = q\vartheta^*(ju) - H.E \qquad (14)$$

$H$ and $E$ represent the coefficient vector, $ju$ represents the current iteration, $q\vartheta^*$ represents the best vector outcomes position, and $q\vartheta$ represents the vector position in the above equation. The coefficient vectors are given by the equation below.

$$F = 2.sm\vartheta, \qquad (15)$$

$$H = 2k.sm\vartheta - k \qquad (16)$$

The random vector between 0 and 1 is represented by $sm\vartheta$ in the following equation. The value of $k$ fell from 2 to 0 during the course of the iteration. The humpback whale bubble net technique may be mathematically expressed using two approaches: spiral updating position and shrinking encircling method. The value of $k$ is lowered in the first technique, as shown in above equation, and the distance between the real position of the whale $(Y, Z)$ and the actual location of the prey $(Y*, Z*)$ is examined in the second way. As a result, the spiral equation was used to represent a helix-shaped group of humpback whales by expressing the real position of the whale and the prey.

$$q\vartheta(ju+1) = \grave{E} \cdot e^{ph} \cdot \cos(2\pi h) + q\vartheta^*(ju) \qquad (17)$$

The distance between the whale and the prey is denoted by the formula $E = |F.q\vartheta^*(ju) - q\vartheta(ju)|$ and is represented as $h$. The constant term is denoted by $p$. To update the solution, the shrinking encircling process is used, which is mathematically stated in further equation,

$$q\vartheta(ju+1) = \begin{cases} q\vartheta^*(ju) - H.E & if\ i < 0.5 \\ E.e^{ph}.\cos(2\pi h) + q\vartheta^*(ju) & if\ i \geq 0.5 \end{cases} \quad (18)$$

$I$ random number ranges from 0 to 1 in the equation above. However, the position vector $qv$ is employed to seek the prey vector, and this vector spans from -1 to 1 to avoid the reference whale by the search agent. The random position vector is denoted as $q\vartheta_{random}$ and is obtained from the preceding solution. As a result, the following is the mathematical formula:

$$E = |F.q\vartheta_{random} - q\vartheta| \quad (19)$$

$$q\vartheta(ju+1) = q\vartheta_{random} - H.E \quad (20)$$

Between estimated output (random hyper-parameter values) and actual output (exact hyper-parameter values), the minimization function is generated. To achieve optimal network structure, the loss function must be reduced. As a result, hyperparameter settings must be adjusted till the achieved error value decreases with each iteration. The WOA optimization technique, which is based on the operating idea of a whale in search of food, is shown here for obtaining optimum weight values in network models. The optimum solution is found by assessing the fitness function. Update solution is conducted based on the fitness function to ensure that the new solution is feasible. The ideal solution is to achieve reduced fitness function. The modified answer is iterated until the maximum number of iterations is reached. The hyper tuning settings are restored in order to get accurate output till the deep learning model's loss function is reduced. As a result, the CNN parameters are improved, and the entire framework is capable of detecting objects in images with single or multiple instances of object.

## 5. Results and Discussion

To validate the suggested technique, a variety of datasets are investigated. The suggested approach is built in Python and compared to other methods like Deep Neural Network (DNN) and Deep Belief Neural Network. Statistical metrics such as accuracy, precision, recall, and F-Measure are used to justify the suggested technique. In the following sections, we will present the overall results as well as comparison graphs of the proposed and existing object detection systems. In this case, 80 percent of the dataset is utilised for training, while the remaining 20% of photos are used for testing.

### 5.1. Performance analysis of proposed model

After conducting the proposed deep learning model in the detection system, its obtained accuracy and loss function are graphically represented in the following figure.

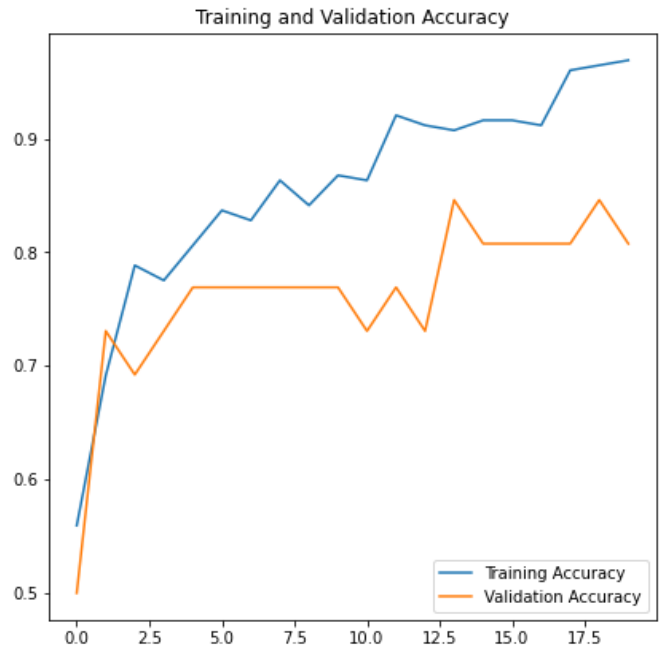

*Figure 2: Simulation outcome of accuracy measure*

Figure 2 depicts the obtained training and validation outcome for accuracy measure. It seems by varying the number of iterations, the obtained accuracy measure gradually increases and keeps on increasing which shows better detection outcome.

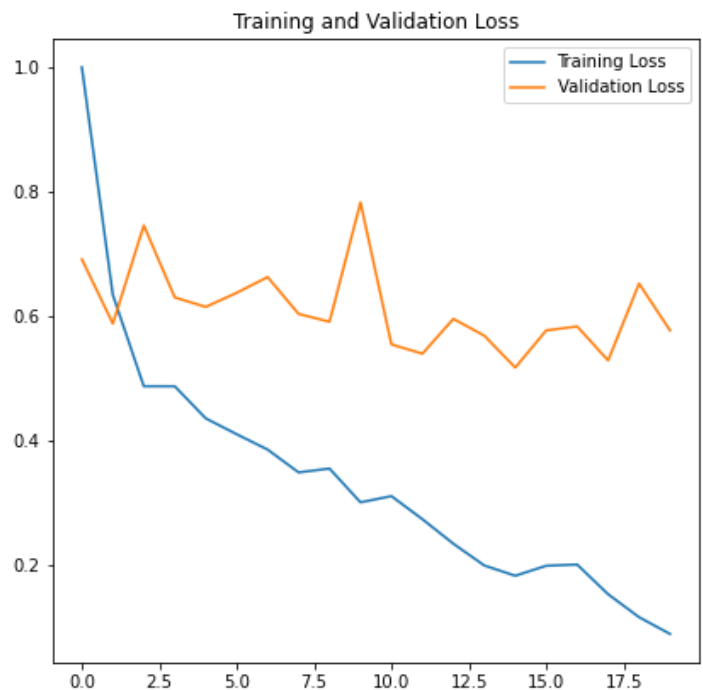

*Figure 3: Outcome of loss function*

Figure 3 furnishes the outcome of loss function which means, prediction error achieved by proposed deep learning technique attains minimum error function additionally, it gets gradually decreased. For instance, when the iteration is increased, the prediction error of the proposed solution decreases. Hence it outperforms better in terms of accuracy and error function. For the clarity of proposed methodology, it is compared with existing techniques and was analysed in the upcoming sections.

## 5.2. Overall Comparison study of proposed and existing techniques

Artificial Neural Network (ANN), Support Vector Machine (SVM), Deep Neural Network (DNN), Deep Belief Neural Network (DBNN), and Pre-trained CNN are compared to the suggested deep learning approach with optimization model.

*Table 1: Performance comparison of proposed and existing prediction system*

| Predictive model | Classifier Performance | | | | |
|---|---|---|---|---|---|
| | Accuracy | Specificity | Sensitivity | Time(s) | Error |
| ANN | 0.826 | 0.8691 | 0.773 | 19.438 | 0.92 |
| SVM | 0.751 | 0.728 | 0.742 | 20 | 0.98 |
| DNN | 0.77 | 0.799 | 0.865 | 34 | 0.86 |
| DBNN | 0.813 | 0.85 | 0.859 | 13.96 | 0.59 |
| Pre-trained CNN | 0.82 | 0.701 | 0.838 | 17.89 | 0.14 |
| **Optimized DCNN** | **0.9864** | **0.9612** | **0.9530** | **9.71** | **0.039** |

The overall comparison of deep learning and machine learning algorithms for the Computer Aided Diagnosis dataset images is shown in Table 2. The error function and processing time are considerably reduced using the suggested strategy. Other metrics such as accuracy, sensitivity, and specificity, on the other hand, are greater for the dataset. For example, the suggested improved DCNN achieves 0.9864 accuracy, while ANN achieves 0.826 accuracy, default CNN model 0.82 accuracy, DBNN achieves 0.813 accuracy, and both DNN and SVM approaches achieve roughly 0.7 accuracy. Similarly, the suggested approach's sensitivity and specificity are 0.9530 and 0.9612, respectively. These two measurements appear to have produced better results than previous strategies.

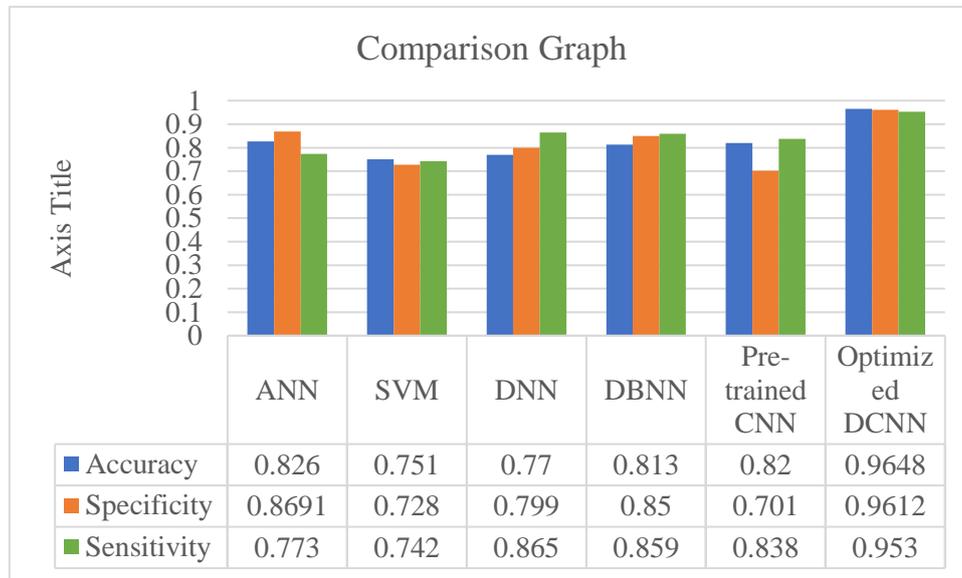

*Figure 4: Comparison of proposed and existing techniques for different metrics*

In terms of accuracy, sensitivity, and specificity, Figure 4 shows the graphical results of suggested and current methodologies. The suggested technique outperforms the others in terms of accuracy, sensitivity, and specificity, demonstrating the framework's viability. The quality of accurately recognising two classes improves as the outcome improves. As a result, in terms of these measures, the suggested strategy produces the most efficient result when compared to other models.

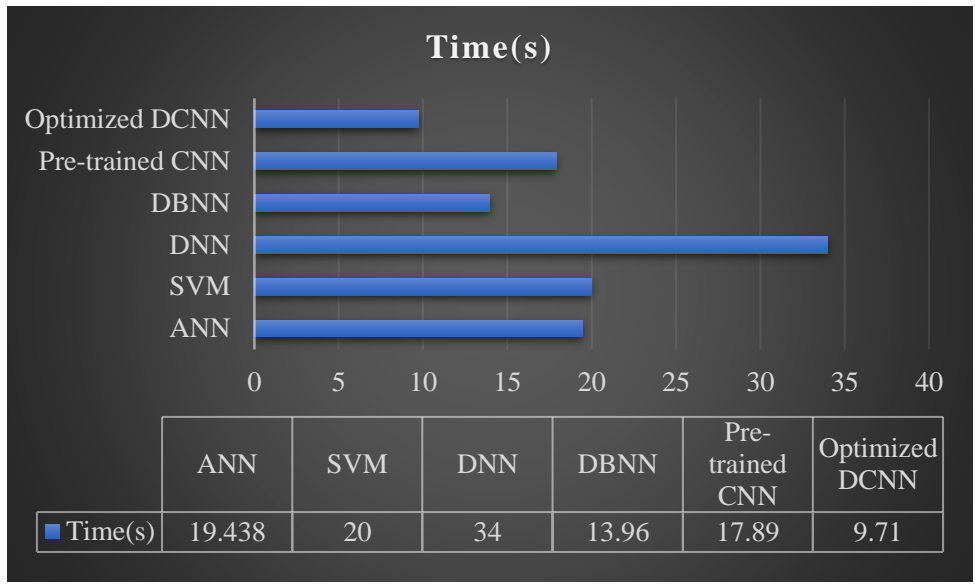

*Figure 5: Overall processing time acquired by proposed and existing approaches*

The entire calculation time required to complete the process using the suggested technique is 9.71 seconds, which is similar to the time required by existing algorithms. As a result, it is evident that the suggested strategy takes substantially less time than alternative methods.

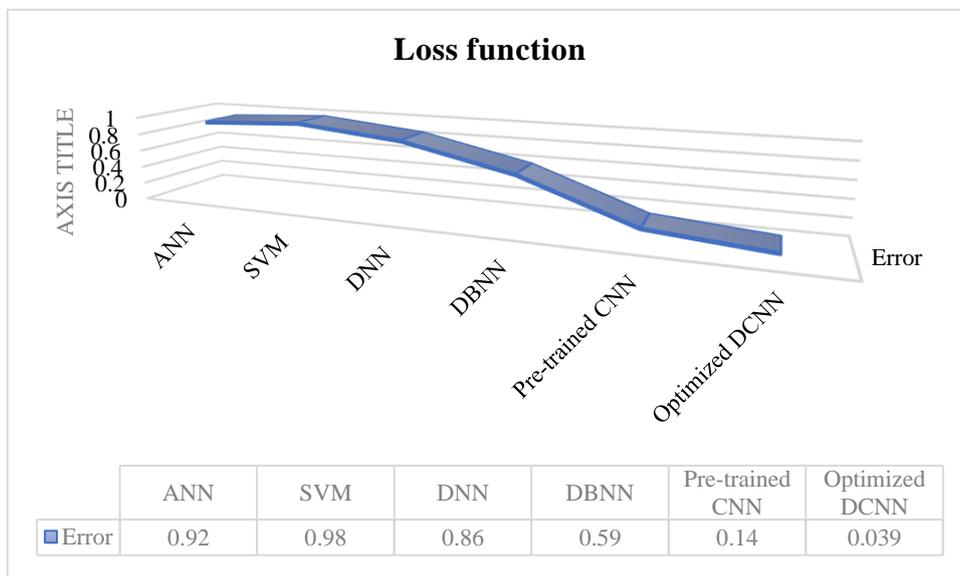

*Figure 6: Error analysis for proposed and existing techniques*

It is a well-known fact that when the error function gained is lower, the system performance increases automatically. By meeting the above fact, the suggested system excels by achieving the lowest possible error value. As a result of the investigation, it can be determined that the object detection model's overall performance has increased sufficiently. The suggested deep learning method achieves a better solution for object detection owing to the insertion of a large number of convolutional layers, according to the overall study.

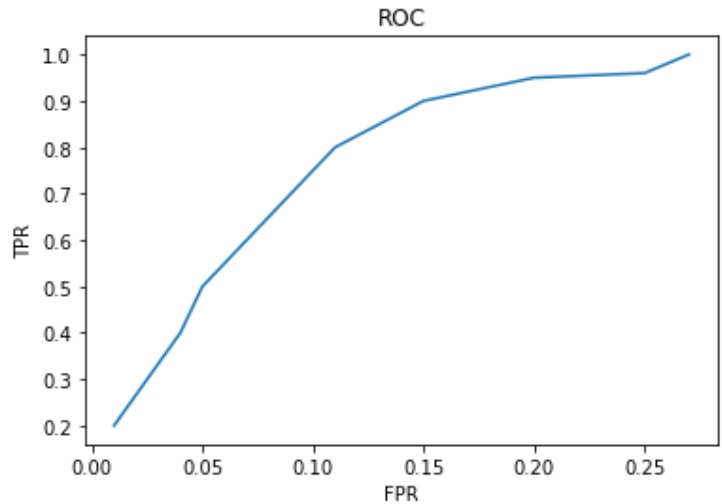

*Figure 7 The ROC Curve*

TPR and FPR are used to map the ROC curve. Figure 8 depicts the suggested DCNN approach's ROC curve. The TPR and the FPR are linked to get a high TPR and a low FPR.

## 6. Conclusion

The purpose of the research is to present an object detection strategy using the proposed hybrid optimized deep learning approach. For the purpose of object detection system, initially with the help of gaussian filter, noisy contents present in an image are eliminated and then background subtraction process is carried to out to determine the prominent region of an image. Then the filtered image is fed into normalization phase to carry out the contrast normalization procedure. The pre-processed outcome is subjected to entropy-based segmentation whereas dissimilar objects are segmented in a single image. Further the segmented image is processed to the object detection phase whereas multiple instance present in an image are detected with the help of proposed hybrid optimized deep convolution neural network. The accuracy acquired by the proposed technique is 0.9864 that is comparatively high than that of the existing methods. The promising results of the proposed method supports its application for object detection in computer visions.